# Compressed Domain Image Classification Using a Dynamic-Rate Neural Network

YIBO XU[1], WEIDI LIU[2], AND KEVIN F. KELLY[1], (Member, IEEE)
[1]Department of Electrical and Computer Engineering, Rice University, Houston, TX 77005, USA
[2]Department of Applied Physics, Rice University, Houston, TX 77005, USA

Corresponding author: Yibo Xu (ybxu2013@gmail.com)

This work was supported in part by the National Science Foundation under Grant CHE1610453, and in part by the Air Force Office of Scientific Research under Grant FA8651-16-C-0185.

**ABSTRACT** Compressed domain image classification performs classification directly on compressive measurements acquired from the single-pixel camera, bypassing the image reconstruction step. It is of great importance for extending high-speed object detection and classification beyond the visible spectrum in a cost-effective manner especially for resource-limited platforms. Previous neural network methods require training a dedicated neural network for each different measurement rate (MR), which is costly in computation and storage. In this work, we develop an efficient training scheme that provides a neural network with dynamic-rate property, where a single neural network is capable of classifying over any MR within the range of interest with a given sensing matrix. This training scheme uses only a few selected MRs for training and the trained neural network is valid over the full range of MRs of interest. We demonstrate the performance of the dynamic-rate neural network on datasets of MNIST, CIFAR-10, Fashion-MNIST, COIL-100, and show that it generates approximately equal performance at each MR as that of a single-rate neural network valid only for one MR. Robustness to noise of the dynamic-rate model is also demonstrated. The dynamic-rate training scheme can be regarded as a general approach compatible with different types of sensing matrices, various neural network architectures, and is a valuable step towards wider adoption of compressive inference techniques and other compressive sensing related tasks via neural networks.

## I. INTRODUCTION

Compressive sensing (CS) [1]–[3] is a mathematical framework for efficient signal acquisition and robust recovery. According to CS theory, a sparse or compressible signal can be reconstructed almost exactly from much fewer number of linear measurements than required by the Nyquist sampling theorem. The CS framework has inspired a variety of imaging systems. There have been comprehensive reviews of current state of the field of CS imaging [4], [5]. Among them, the single-pixel camera (SPC) [6] is especially useful in imaging outside of the visible regime, e.g. infrared imaging [7], terahertz imaging [8] and hyperspectral imaging [9] where detector arrays are extremely expensive or even nonexistent. In many applications, the goal is to solve an inference problem like anomaly detection or classification, instead of reconstructing the full image. In a traditional camera system, such tasks rely on acquiring high quality images and extracting features from the images for inference [10]–[12]. However, in a CS imaging system, full image reconstruction from compressive measurements poses great challenges for applications on resource-limited platforms because of drawbacks in CS image reconstruction methods: traditional iterative optimization algorithms [13] are computation-heavy and time-consuming, especially for high dimensional signals such as large images, hyperspectral and video data; the recently deep learning approaches [14] require large amount of training data, and the performance is highly dataset-dependent and typically do not scale well to high dimensional signals. Therefore, for a lot of real-world applications outside the visible regime where SPC is preferably used for sensing, bypassing the image reconstruction step would be highly beneficial, e.g. for high-speed object recognition and in resource-constraint platforms, where people try to drive down the number of measurements, computation complexity, inference time and power consumption.

  

In this paper, we focus on using neural networks to perform image classification directly on CS measurements and to push the limit of object detection and classification under very limited resources both on sensing side and processing side. There has been a lot of research in compressed domain inference [15]–[20]. Davenport et al. [19] employed the matched filter of compressed sensing patterns applied to a library of images to create a 'smashed' filter and demonstrated the validity of the random projections-based approach for compressed domain image classification. Later, Li et al. employed the same SPC system but used learned sensing patterns through data-dependent ''secant projections'' for the same task [20]. Recently, the convolutional neural network (CNN) has been employed [16], [17] and produced much higher classification accuracy compared to other approaches.

However, current neural network methods require training a separate network model for each different measurement rate. The ratio of the number of linear measurements to the number of pixels in the reconstructed image is called the measurement rate (MR). For a variety of applications, the MR would keep changing or is indefinite in a compressive imager, leading to the need of training of a large number of neural network models which is inefficient in storage and computation. In addition, the classification accuracy remains low at very low MR like 0.01, but the low MR is a regime that is especially interesting and useful.

In this paper, we directly address these limitations. The contributions of the paper are as follows.

To begin with, we develop a general approach called CS dynamic-rate neural network (DRNN) training scheme that allows a single trained neural network to perform compressed domain image classification over a range of MRs using fixed sensing patterns. The effectiveness of this approach is demonstrated on datasets including MNIST, CIFAR-10, Fashion-MNIST and COIL-100, and the DRNN produces approximately equal classification accuracy at each MR as that of a single-rate neural network valid only for that MR. The variety of datasets shows that DRNN can be used for images of different scales. Most importantly, the network is trained on CS measurements of a few different MRs within the range of interest instead of being trained on all different MRs, leading to great savings in computation and power consumption. Such training scheme can also be regarded as opening up an extra dimension for performing data augmentation that uniquely exists in CS related tasks to provide the neural network with dynamic-rate property. Next, we perform the DRNN training scheme using various network designs, including the simple 2-layer feedforward neural network (2L-FFNN) and the deeper convolutional neural networks (CNNs). Its success shows that the DRNN training scheme can be used for both simple and complex neural network architectures. Lastly, two types of fixed binary sensing patterns are studied. In particular, the relatively new binary Partial Complete (PC) matrix [21], [22] which is a structured matrix is compared with the permuted Walsh-Hadamard (PWH) matrix which in effect is a random matrix. Results show that PC matrix significantly increases the classification accuracy at very low MRs like 0.01, which we posit is due to the low frequency patterns from PC matrix picking up more information of the image.

## II. BACKGROUND AND RELATED WORK
### A. COMPRESSIVE SENSING

Compressive sensing seeks to minimize the number of measurements to be taken from a signal while still retaining the information necessary to produce a nearly complete recovery, exploiting the structure of the signal. A core result of CS states the following: a length $N$ signal (or image) $x$ that is $K$-sparse in an arbitrary basis can be recovered from $M = O(Klog(N/K))$ linear measurements. Formally, a signal $x = \Psi\alpha$ where $\Psi$ is an orthonormal basis of size $N \times N$, and $\alpha$ is $K$-sparse vector of size $N \times 1$, can be exactly reconstructed from measurements

$$y = \Psi x = \Phi\Psi\alpha \quad (1)$$

where $\Phi$ is an $M \times N$ matrix called the sensing matrix or sensing patterns, provided that the product matrix $\Phi\Psi$ satisfies a certain restricted isometry property (RIP) [21]. The ratio of $M/N$ is the measurement rate (MR). The measurement $y$ is a vector of weighted sum of entries in signal $x$, representing a highly concise encoding of $x$, based on which the image inference can be made. In real world experiments, the measurement $y$ is usually corrupted by noise represented by vector $\epsilon$, so the measurement $y$ can be represented as

$$y = \Psi x + \epsilon = \Phi\Psi\alpha + \epsilon \quad (2)$$

To recover the signal $x$ from the random measurement vector $y$, it has been shown that $l_1$ optimization could be used [1], [2], [23]:

$$\hat{\alpha} = argmin \|\alpha\|_1 \quad s.t. \quad \|y - \Phi\Psi\alpha\|_2 < \epsilon \quad (3)$$

Total variation (TV) regularization [24] is another well-known method for its ability to recover the edges or boundaries more accurately than $l_1$ method:

$$\hat{x} = argmin \sum_i \|D_i x\| \quad s.t. \quad \|y - \Phi x\|_2 < \epsilon \quad (4)$$

where $D_i x$ is the discrete spatial gradient vector of the image $x$ at pixel $i$, and $\|D_i x\|$ is the $l_2$ norm of $D_i x$.

However, signal reconstruction through iterative optimization algorithms is computationally expensive and time consuming due to the algorithm being iterative in nature and not suitable for parallelization. Furthermore, the parameters in the algorithm need to be tuned or defined by the user making the algorithm difficult for practical use. The deep learning approaches for CS signal reconstruction, though much faster, require intensive and prolonged training process and only perform well on datasets that are similar to the training dataset, and need to be retrained or fine-tuned if to be used with different types of image data.

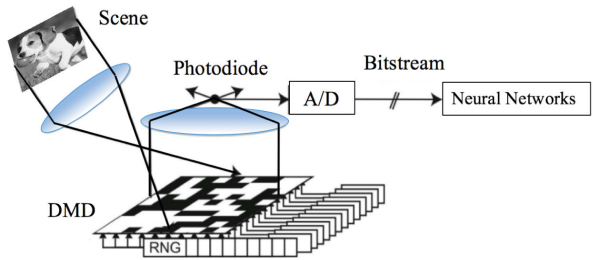

**FIGURE 1.** Schematic of compressed domain image classification through neural networks. CS measurements of the scene are acquired by the SPC implementing the sensing matrix. Each row of the sensing matrix is reshaped into a 2D sensing pattern and displayed on the DMD over time while the CS measurement of the scene corresponding to that row is acquired by the photodiode. The CS measurements are fed into the trained neural network for classification, bypassing reconstruction step.

### B. SINGLE-PIXEL COMPRESSIVE IMAGING

Whereas this paper is simulation only, the final result is applicable to a variety of compressive imaging systems. Here we focus on its implementation in the original single-pixel camera architecture (SPC) [6]. The SPC, as shown in Fig. 1, is an imaging platform that implements the image encoding process specified by (1) in the optical domain. The scene, which is the original signal $x$, is focused by a convex lens onto a digital micromirror device (DMD) functioning as a spatial light modulator. The DMD consists of an array of micro-sized mirrors where each mirror can tilt at either $+12°$ or $-12°$ about their diagonals. To encode the signal, the DMD is programmed to display over time a sequence of sensing patterns which modulate the intensity of each scene pixel. Each sensing pattern comes from one row of the sensing matrix reshaped into a 2D configuration. When the image of the scene is projected onto the DMD, in effect the DMD mirrors tilting $+12°$ (ON state) encodes 1 and the mirrors tilting $-12°$ (OFF state) encodes 0 on the image pixels. The light reflected by $+12°$ mirrors comes out in one direction and by $-12°$ mirrors comes out in an opposing direction, and a second lens sums up the light coming in the $+12°$ direction while the rest of the light is discarded. The summed up light is detected by a single-pixel photon detector as measurement data corresponding to the current sensing pattern displayed on DMD. The standard type of sensing matrix used in the original SPC is random binary matrix such as permuted Walsh-Hadamard (PWH) matrix.

Although existing SPC architectures allow for displaying floating-point values by temporal modulation of the micromirrors being in the ON state during each measurement period, floating-point sensing patterns take up more memory and lead to reduced pattern speed compared to binary patterns and therefore not preferable in real applications.

### C. COMPRESSED DOMAIN IMAGE CLASSIFICATION

In regards to compressed domain inference, Calderbank *et al.* [25] provided the first theoretical results that learning directly in the compressed domain is feasible. In particular, they provided bounds demonstrating that the performance of a linear SVM in the compressed domain is close to the performance of the best linear classifier in the uncompressed domain and that classifiers can be learned directly in the compressed domain. Davenport *et al.* [19] employed the 'smashed' filter where the random sensing matrix and a 1-nearest-neighbor classifier were used. Later, the same SPC system was employed but with learned patterns through data-dependent "secant projections" to perform classification directly in the compressed domain [15]. Most recently, Lohit *et al.* [16] employed a convolutional neural network (CNN) for compressed domain classification, which produced much higher classification accuracy compared to the smashed-filtering approach as well as being computationally efficient. Following this work, Adler *et al.* [17] developed an end-to-end deep learning solution for compressed domain classification where sensing matrix is jointly learned with the inference operator. In the previous two cases, CS measurements are first projected to the image space to get $\tilde{x} = \Phi^T y$, or $\tilde{x} = \Phi^+ y$, or $\tilde{x} = \tilde{\Phi} y$, where $\Phi^T$ is the transpose of sensing matrix $\Phi$, $\Phi^+$ is pseudo-inverse of $\Phi$, $\tilde{\Phi}$ is the learned matrix from weights in a fully connected layer, and $\tilde{x}$ can be regarded as a proxy of $x$ with the same dimension as $x$.

Recently, Prof. Pavan Turaga and his research group developed a rate-adaptive neural network for compressed domain classification [26]. As far as we know, the proposed DRNN algorithm and the rate-adaptive method [26] are the only two research works that attempted to use a single network to handle varying MRs, and three main aspects differentiate the two algorithms: scope of application, principles of the algorithm, and computation complexity. To begin with, DRNN algorithm is generally applicable whenever a sensing matrix is given beforehand, whether it is a random matrix, structured matrix, or any learned sensing matrix; whereas the rate-adaptive network learns the sensing matrix and is not applicable when one wants to use any given sensing matrix, whether it is learned or not. So the scope of application does not overlap between the two algorithms. Next, the idea behind how to make the network valid for multiple MRs are fundamentally different. The weights of the trained DRNN are different from any single-rate networks, because measurements of multiple MRs are in training data which actually impact the network weights. For the rate-adaptive method, however, after the inference network was first trained at the minimal MR, its weights are freezed forever. In order to work with multiple MRs, the algorithm adds a new neuron each time in the first fully-connected layer and only train this new neuron with the previously trained weights part fixed, and in this way, learn a new sensing pattern that is optimized for the fixed inference network. Thus, technically speaking, the rate-adaptive method trains a single-rate inference network first, then "adapts" new sensing patterns to the fixed inference network. It is more like a method for learning the sensing matrix which is optimized for the dataset and the single-rate inference network. Lastly, the proposed DRNN algorithm is more efficient during training because it can use a small number of MRs in training data instead of all MRs. The rate-adaptive method, on the other hand, needs to train the network for a whole cycle to learn each new sensing pattern. Apart from the above works, efforts have also been made to jointly learn binarized sensing patterns together with the inference part using neural networks to facilitate binary pattern display on the DMD [27].

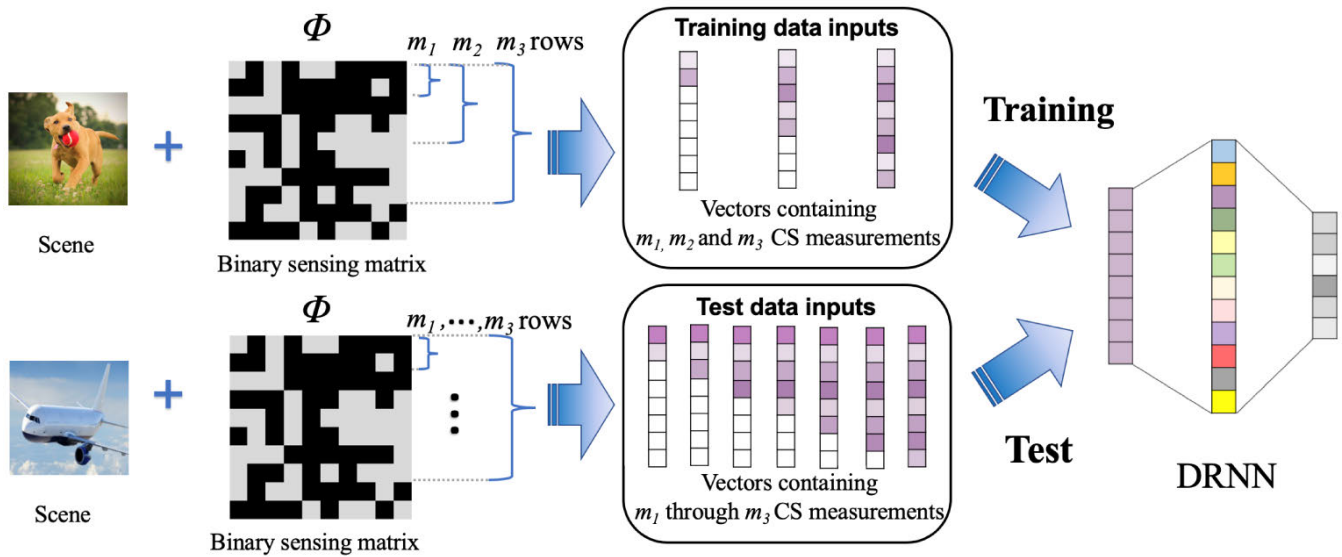

**FIGURE 2.** DRNN training and test scheme when using a 2L-FFNN. CS measurements of the scene are acquired by SPC implementing different number of rows of the sensing matrix $\Phi$. In this schematic, $m_1 = m_{min}$ and $m_3 = m_{max}$ denoting the range of CS measurements chosen for the DRNN to perform classification with. For training the DRNN, a few intermediate MRs are selected and the corresponding CS measurement vectors for every training image are used in training. For simplicity of illustration, only one intermediate number denoted by $m_2$ is shown in the schematic. Measurement vectors with a smaller number of CS measurement entries than $m_3$ are padded with zeros so they are all of length $m_3$. The trained DRNN can classify on $m_1$ all through $m_3$ CS measurements of test images.

## III. CS DYNAMIC-RATE NEURAL NETWORKS

As elaborated in the introduction section, there are various limitations associated with previous neural network methods. This section first describes our CS dynamic-rate neural network (DRNN) training scheme for different types of neural network architectures, point out its benefits, and then discuss the advantages of using fixed binary sensing patterns and of using PC sensing matrix to raise the classification accuracy at very low MRs.

### A. CS DRNN TRAINING SCHEME

We describe our general approach of CS DRNN training scheme that enables a single trained neural network to classify over a range of chosen MRs of interest. Fig. 2 demonstrates the DRNN training scheme when using the 2L-FFNN. In contrast, we refer to a neural network trained in the traditional way which is valid for only one MR as a single-rate neural network. Simulation in this paper shows that the DRNN produces approximately equal performance at each MR as that of a single-rate neural network for the particular MR.

To train a DRNN, first we choose the desired range of MRs that the DRNN is expected to work with, and use $mr_{min}$ and $mr_{max}$ to denote the minimum MR and maximum MR. The corresponding number of CS measurements are $m_{min}$ and $m_{max}$. For training, a set of intermediate MRs between $mr_{min}$ and $mr_{max}$ are selected: $\{mr_1, mr_2, \ldots, mr_k\}$, where $k$ is the total number of selected intermediate MRs. For every training image, CS measurements of the selected MRs are generated.

These measurement vectors have different lengths that will be inputs to the neural network with a fixed input layer size. We develop two approaches to solve the size mismatch between the CS measurement vectors and input layer. Two types of neural networks are experimented in the paper: the 2L-FFNN and the CNN. In the case of the 2L-FFNN, the CS measurement vectors are zero-padded at the end so they are all of length $m_{max}$, which is also the dimension of the input layer of network. All these zero-padded CS measurement vectors of all the intermediate MRs constitutes the new final training dataset. In the case of the CNN, the CS measurement vector $y$ is first projected back into the image space dimension to get $\tilde{x}$ by $\tilde{x} = \Phi^+ y$, where $\Phi^+$ is the pseudo-inverse of the sensing matrix $\Phi$, and $\tilde{x}$ is a proxy of the original image $x$ with the same dimension as $x$, which is also the size of the input layer of the CNN. These $\tilde{x}$ corresponding to all the intermediate MRs for all the images in the training dataset constitutes the new final training dataset.

Then the network is trained on the new final training set with standard optimization algorithms like Adam [28]. It is crucial to randomly shuffle the data points in the training set. Training on a series of data batches where each batch is of a single MR fail to work as we have learned through experiments.

A key question in the DRNN training scheme is: how large should $k$ be? The naive approach where all MRs between $mr_{min}$ and $mr_{max}$ are selected, however, will cause the final training set to explode and drastically increase the training time and complexity if not infeasible at all. And we have found that it is unnecessary. As is shown in the experiments in this paper, training the network on only a few of the intermediate MRs is sufficient to achieve expected performance on all MRs between $mr_{min}$ and $mr_{max}$ and that adding more MRs for training does not show significant improvement.

The CS DRNN training scheme is simple in procedure and efficient in computation. It involves minimal processing preparing the new training set and uses standard neural network training algorithms with on need for modification or fine-tuning. The DRNN training scheme bears similarities but also key differences compared to the conventional data augmentation methods used for preprocessing full images such as cropping, flipping and rotation. In training DRNN, $m_{min}$ can be a small fraction of $m_{max}$, e.g. $m_{min}$ is approximately 4% of $m_{max}$ in our simulation. The CS DRNN training scheme can be regarded as opening up another dimension for performing data augmentation that uniquely exists in CS related tasks to provide the neural network with dynamic-rate property, and it is in parallel to the conventional method. In this sense, two types of data augmentation can be performed for compressed domain classification using neural networks: the conventional image augmentation which makes the system invariant under changes due to object location, lighting, angle, etc., and CS dynamic-rate data augmentation which endows the system with dynamic-rate property.

### B. PARTIAL COMPLETE SENSING MATRIX

In this paper, we focus on using fixed binary sensing matrices for taking CS measurements in contrast to random Gaussian [16] or learned sensing matrix [17] which are in full-precision.

We study two types of binary sensing patterns: the randomly permuted Walsh-Hadamard (PWH) matrix, and the Partial Complete (PC) matrix [21], [22]. PWH is generated by randomly permute the rows and columns, of the Walsh-Hadamard matrix. In the PC matrix, the rows are grouped into blocks where the same signature pattern is shared within a block of rows, and the measurement values for rows in each block tend to have similar intensities. The first block of rows in the PC matrix consists of low frequency patterns, and since most natural images have energy concentrated in low frequency region, we expect PC matrix to perform better at very low MRs than PWH matrix.

Fixed binary sensing patterns are more beneficial in practice than full-precision random Gaussian or learned patterns. This is because, even though displaying floating-point patterns on the DMD in a SPC platform is possible by temporal modulation of bit planes of the sensing pattern, it leads to reduced pattern speed and increased storage cost in memory compared to binary patterns. Also, fixed binary sensing patterns are dataset-independent and do not require learning from full image datasets which might not exist at all or extremely expensive to acquire, like infrared or hyperspectral datasets.

## IV. EXPERIMENTAL RESULTS

In this section, we present the results of performing the CS DRNN training scheme using the 2-layer feedforward neural network (2L-FFNN) and the CNN. We also compare the PC matrix and PWH matrix for sensing and found that PC sensing matrix greatly enhances the classification accuracy at very low MRs.

### A. DATASETS

We perform the DRNN training scheme on four widely used datasets: MNIST [29], CIFAR-10 [30], Fashion-MNIST [31] and COIL-100 [32]. We slightly modified them to suit our problem setting. Example images of each modified dataset are shown in Fig. 3. MNIST and Fashion-MNIST each consists of 28 × 28 grayscale images belonging to 10 categories, with 60000 training images and 10000 test images. Since PWH matrix and PC matrix of dimension 784 × 784 required for sensing 28 × 28 images do not exist, we pad zeros around four sides of every image evenly so that the final image size becomes 32 × 32, and use 1024 × 1024 PWH and PC matrices for sensing. The CIFAR-10 dataset contains 32 × 32 color images of 10 classes with 50000 training images and 10000 test images. Since our problem setting of interest concerns a monochromatic single-pixel camera with a single photodetector for short-wave infrared (SWIR) where CS measurements are by nature grayscale, we use only use the intensity channel of CIFAR-10 images for classification to take this into account, which could reduce classification accuracy but more accurately reflects the type of SWIR image data that would be acquired. The COIL-100 dataset contains color images of 100 objects of size 128 × 128. The objects were placed on a motorized turntable against a black background and images were taken at pose internals of 5 degrees. We use only the intensity channel of COIL-100 images for 100 object recognition with the DRNN in the paper.

### B. TRAINING DETAILS

For the DRNN in this paper, we choose $mr_{min} = 0.01$, $mr_{max} = 0.25$ as an example because we want to include very low MRs like 0.01 which is a challenging case and of great importance in resource-constrained platforms. We study the DRNN performance of using different numbers of intermediate MRs. For simplicity, we use '$k$-MR training' to indicate that $k$ different MRs are used for training for every image in the original training image dataset. The trained neural network is tested on CS measurement vectors containing $m_{min}$ all through $m_{max}$ measurements.

For MNIST and CIFAR-10, we compare training with 4, 6, 10 and 50 intermediate MRs using the 2L-FFNN, and compare between PWH and PC sensing matrices. The 2L-FFNN has a fully-connected layer with 400 neurons after the input layer.The same trend is observed for MNIST and CIFAR-10 that 10-MR training is sufficient to achieve expected performance. For Fashion-MNIST and COIL-100, 5-MR and 10-MR training results using PWH sensing matrix are shown and compared with single-rate neural networks. For Fashion-MNIST, the network design is a CNN with 3 convolutional layers each followed by a max-pooling layer. For COIL-100, a CNN with 6 convolutional layers each followed by a max-pooling layer is used. The COIL-100 dataset is randomly split into 30 % and 70 % for training and testing, respectively.

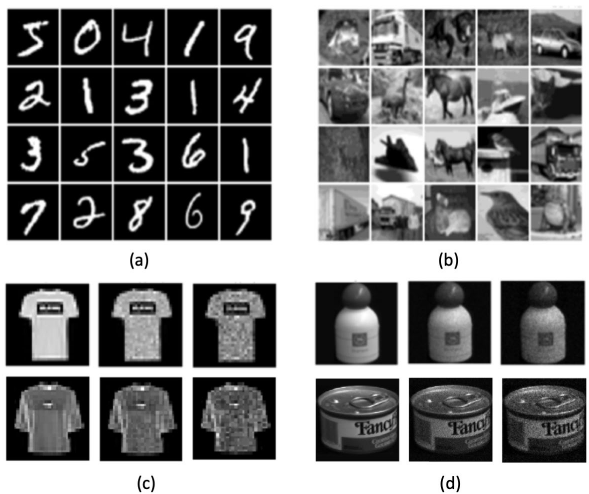

**FIGURE 3.** Example images of the datasets and with added noise: (a) MNIST dataset, (b) grayscale CIFAR-10 dataset, (c) Fashion-MNIST dataset, (d) Grayscale COIL-100 dataset. In (c) and (d), within each row, from left to right: clean image, with Poisson noise of n = 200, with Poisson noise of n = 50.

For zero-padded MNIST, CIFAR-10, and zero-padded Fashion-MNIST, which has image size of 32 × 32, the numbers of CS measurements selected for training are: for 4-MR training, $m = [10, 51, 102, 256]$, for 6-MR training, $m = [10, 20, 51, 102, 150, 256]$; for 10-MR training, $m = [10, 18, 26, 34, 42, 51, 75, 102, 180, 256]$; for 50-MR training, $m$ starts from 10 with an increment of 5 until 250, plus 256. For COIL-100 which has 128 × 128 images, the same corresponding MRs are selected. We follow the general principle in selecting $m$ that more MRs are selected in the low MR region where the network classification performance changes rapidly, and less in high MR region where performance changes slower. The loss function used is cross-entropy. We use the Adam [28] optimizer for training with the following parameters: initialLearningRate = 5e-5, learningRateDropFactor = 0.9, learningRateDropPeriod = 4, total number of epochs = 100.

### C. RESULTS AND DSICUSSION

Fig. 4 and Fig. 5 show the test accuracy of compressed domain image classification on zero-padded MNIST for the dynamic-rate 2L-FFNN using PWH sensing matrix and PC sensing matrix, respectively. Comparison is shown among different DRNNs that are trained with 4, 6, 10 and 50 different intermediate MRs, respectively. The single-rate network is trained with one MR of 0.25 corresponding to 256 measurements and is expected to perform well only for predicting 256 measurements. In the figures, each curve represents the test result for one neural network. The MRs used in training are represented by markers on each accuracy curve.

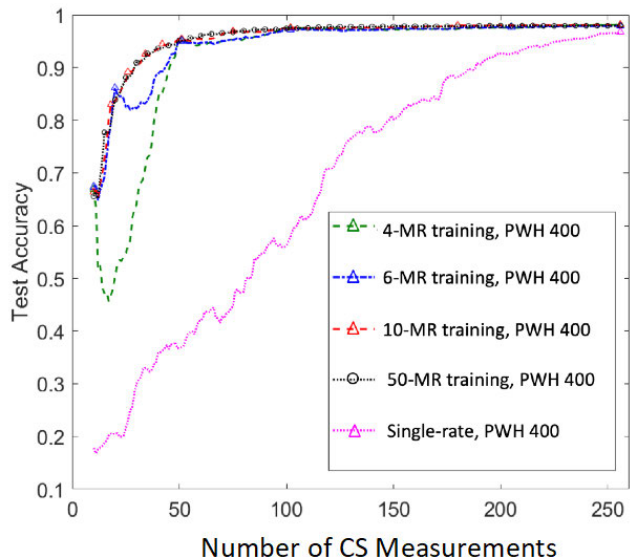

**FIGURE 4.** Test accuracy comparison of DRNNs of different k-MR training and the single-rate neural network on zero-padded MNIST dataset using Permuted Walsh-Hadamard (PWH) sensing matrix. The network is the 2L-FFNN with 400 neurons in the fully-connected layer. Comparison is shown among DRNNs that are trained with 4, 6, 10 and 50 different intermediate MRs between 0.01 and 0.25. The single-rate network is trained only with MR = 0.25 corresponding to 256 measurements. The selected MRs used for training are indicated by markers on the curves.

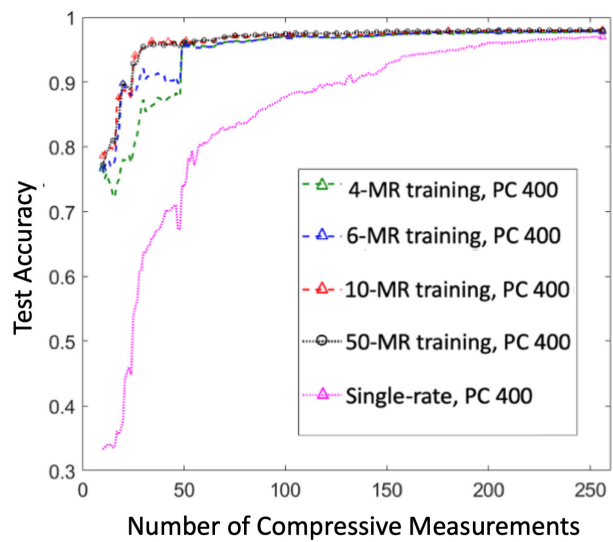

**FIGURE 5.** Test accuracy comparison of DRNNs of different k-MR training and the single-rate neural network on zero-padded MNIST dataset using Partial Complete (PC) sensing matrix. The network is the 2L-FFNN with 400 neurons in the fully-connected layer. Comparison is shown among DRNNs that are trained with 4, 6, 10 and 50 different intermediate MRs between 0.01 and 0.25. The single-rate network is trained only with MR = 0.25 corresponding to 256 measurements. The selected MRs used for training are indicated by markers on the curves.

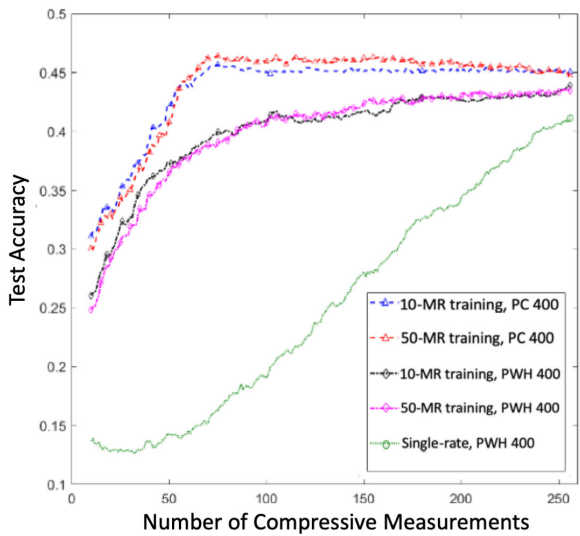

**FIGURE 6.** Test accuracy for different DRNNs and single-rate neural networks on grayscale CIFAR-10. The network design is the 2L-FFNN with 400 neurons in the fully-connected layer. Comparison is shown among DRNNs trained with 10-MR and 50-MR training using PC and PWH sensing matrices. The single-rate network is trained with only one MR of 0.25 corresponding to 256 measurements.

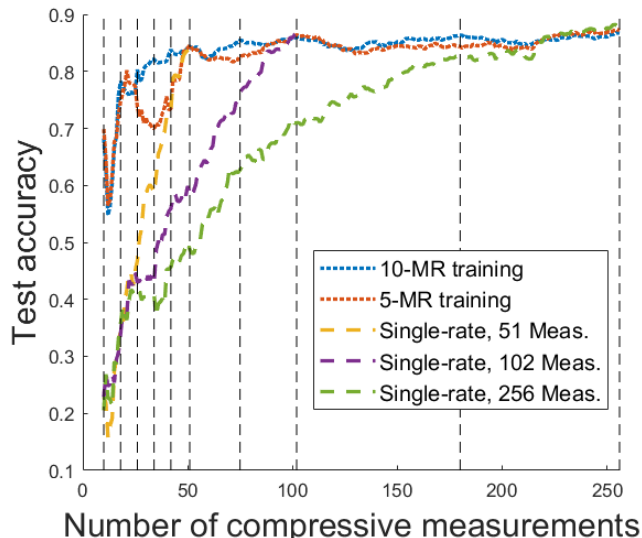

**FIGURE 7.** Test accuracy for different DRNNs and single-rate neural networks on Fashion-MNIST dataset. All the DRNNs and single-rate networks are trained using the PWH sensing matrix and a CNN with 3 convolutional layers. 5-MR training and 10-MR training results are shown. The three single-rate networks are trained at 51, 102, and 256 measurements, respectively, and tested on a range of number of measurements.

It is clearly demonstrated that, for both PWH matrix and PC matrix, 10-MR training is sufficient. 50-MR training does not significantly improve the results. For either 4-MR or 6-MR training, the accuracy curves have large dips for MRs that did not get selected in training especially at very low MRs. Also the single-rate network performs poorly compared to DRNNs at MRs that it was not trained on. Besides, the PC matrix has much better accuracy than PWH matrix at low MRs.

Results for grayscale CIFAR-10 are shown in Fig. 6. Same as MNIST, 10-MR training produces approximately equal performance as 50-MR training and the single-rate model has poor performance compared to DRNNs. We show only the 10-MR and 50-MR training results for grayscale CIFAR-10 since we found that the trend is exactly the same as with MNIST for 4-MR and 6-MR training which generates large dips on the curve. It is clearly shown that PC sensing matrix has much higher test accuracy at very low MRs compared to PWH matrix for both MNIST and grayscale CIFAR-10. This could be because that the first 64 sensing patterns from the 1024 × 1024 PC matrix are all low frequency patterns and therefore capture the core and important information of the image to facilitate classification. Then some of the following sensing patterns from the PC matrix are high frequency patterns, which could capture nuance of the image leading to a slight drop of performance with more measurements. We have also observed the following: while the accuracy curve is smooth for 10-MR and 50-MR training using the PWH matrix, the PC matrix curves have step-like features. It could be because that the PWH is a random matrix and each measurement picks up about the same amount of information of the image to facilitate classification, while the measurement values using PC matrix could fluctuate a lot contributing different amount of information.

We have noticed that in [26], CIFAR-10 dataset has accuracy around 0.7 through jointly learning the sensing matrix and inference network, which is higher than in our paper. The accuracy difference is largely due to deeper network architecture used in [26]. If not using a deep enough network, neither the learned sensing matrix nor the random matrix can achieve 0.7 accuracy, no matter how many measurements used. With an appropriate deep network, the learned sensing matrix may need a smaller number of measurements, because it captures more information specific to the dataset and task. The random matrix is a universal encoding scheme without the need for learning. So, with an appropriate deep network model, the non-adaptive random sensing matrix used in the paper can also achieve around 0.7 accuracy or higher when more measurements are used.

Given the validation and success of the DRNN training scheme, we then also compare the performance between the DRNN at a given MR with the performance of a single-rate neural network for that particular MR. We train single-rate neural networks valid for MRs 0.01, 0.05, 0.1 and 0.25 separately and compare with 10-MR DRNN training. For MNIST, we also test the method in [16] for single-rate networks where we use Gaussian random sensing matrix and the Lenet-like neural network with two convolutional layers using $\tilde{x} = \Phi^T y$ as input. Test results are also compared at MRs of 0.08 and 0.15 which none of the DRNNs or single-rate networks were trained on. Results are shown in Table 1. There are two accuracies in each table cell for MRs 0.15 and 0.08 for single-rate neural networks, where the first is predicted by the single-rate network below this cell, and the second is predicted by the single-rate network above this cell. For grayscale CIFAR-10, as a comparison with the dynamic-rate 2L-FFNN, we also tried single-rate networks using $\tilde{x} = \Phi^+ y$ as input where $\Phi^+$ is the pseudo-inverse of $\Phi$, and tested on a network with 3 convolutional layers and another network with 7 residual layers. Results for grayscale CIFAR-10 are shown in Table 2. For both datasets, it is shown that a single dynamic-rate 2L-FFNN produces approximately equal performance as single-rate models using 2L-FFNN and using deeper networks with convolutional and residual layers at the trained MRs of 0.01, 0.05, 0.1 and 0.25. Also, PC matrix increases classification accuracy compared to PWH matrix at very low MRs of 0.01 and 0.05.

**TABLE 1.** Test accuracy for different DRNNs and single-rate neural networks on grayscale CIFAR-10. The network design is the 2L-FFNN with 400 neurons in the fully-connected layer. Comparison is shown among DRNNs trained with 10-MR and 50-MR training using PC and PWH sensing matrices. The single-rate network is trained with only one MR of 0.25 corresponding to 256 measurements.

| MR | Number of CS Measurements | DRNN, PC, 2L-FFNN 400, 10-MR training | DRNN, PWH, 2L-FFNN 400, 10-MR training | Single-rate, PC, 2L-FFNN 400 | Single-rate, PWH, 2L-FFNN 400 | Single-rate, Gaussian, 2 convolutional layers network |
|---|---|---|---|---|---|---|
| 0.25 | 256 | 98.05% | 98.08% | 96.93% | 96.90% | 98.06% |
| *0.15* | *154* | *97.53%* | *97.71%* | *96.58% / 93.33%* | *96.45% / 85.28%* | *68.13% / 70.17%* |
| 0.1 | 102 | 97.19% | 97.54% | 96.58% | 96.45% | 96.68% |
| *0.08* | *82* | *96.93%* | *96.73%* | *95.46% / 94.87%* | *92.34% / 91.43%* | *50.14% / 62.60%* |
| 0.05 | 51 | 96.31% | 95.41% | 95.46% | 92.34% | 94.08% |
| 0.01 | 10 | 78.61% | 66.43% | 79.60% | 69.31% | 68.24% |

**TABLE 2.** Test accuracy for different DRNNs and single-rate neural networks on grayscale CIFAR-10. The network design is the 2L-FFNN with 400 hidden neurons. Comparison is shown among DRNNs trained with 10-MR and 50-MR training using PC and PWH sensing matrices. The single-rate network is trained with only one MR of 0.25 corresponding to 256 measurements.

| MR | Number of CS Meas. | DRNN, PC, 2L-FFNN, 10-MR training | DRNN, PWH, 2L-FFNN, 10-MR training | Single-rate, PC, 2L-FFNN | Single-rate, PWH, 2L-FFNN | Single-rate, PWH, pseudo-inverse projection, 3 convolutional layers | Single-rate, PWH, pseudo-inverse projection, 7 residual layers |
|---|---|---|---|---|---|---|---|
| 0.25 | 256 | 45.08% | 43.89% | 41.38% | 41.13% | 44.95% | 46.05% |
| *0.15* | *154* | *45.28%* | *41.46%* | *42.62% / 35.61%* | *39.69% / 28.32%* | *36.02% / 38.49%* | *34.61% / 34.81%* |
| 0.1 | 102 | 44.93% | 41.36% | 42.62% | 39.69% | 41.40% | 39.86% |
| *0.08* | *82* | *45.52%* | *39.95%* | *39.81% / 40.48%* | *38.72% / 36.19%* | *29.69% / 37.35%* | *25.57% / 36.29%* |
| 0.05 | 51 | 42.31% | 37.28% | 39.81% | 38.72% | 37.27% | 35.61% |
| 0.01 | 10 | 31.10% | 26.02% | 33.86% | 26.87% | 26.49% | 24.89% |

To demonstrate the effectiveness of the DRNN training scheme on more complex neural network architectures and on images of larger scales, results for Fashion-MNIST and COIL-100 datasets are shown in Fig. 7 and Fig. 8, respectively. The same trend in the results of using different number of MRs is observed as for MNIST and CIFAR-10, so we are showing only the 5-MR and 10-MR results here. Results for different single-rate networks using the same network design are also shown for Fashion-MNIST.

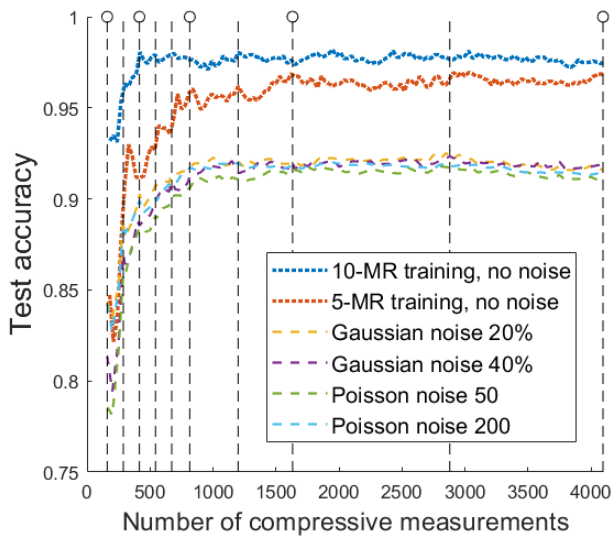

**FIGURE 8.** Test accuracy for DRNNs on the grayscale COIL-100 dataset. Comparison of DRNNs between 5-MR and 10-MR training with no added noise are demonstrated. Noise performance are shown of DRNNs of 10-MR training at added Gaussian noise levels of 20% and 40% and Poisson noise of $n = 50$ and $n = 200$.

Besides image classification, the proposed DRNN training scheme can be applicable in many other CS related tasks. Compressive sensing magnetic resonance imaging (CS-MRI) is an emerging and popular field which aims to reconstruct MRI images from compressive measurements with the benefits of greatly reducing the time of scanning a patient. Deep learning has been successfully used for CS-MRI [33]–[35]. Currently, a different neural network model needs to be trained for each different MR. The proposed idea and method of DRNN training can be generalized and applied to CS-MRI with appropriate modifications, which is a topic worth further exploration and research.

## V. ROBUSTNESS TO NOISE

In real world applications, a significant factor affecting the performance of the proposed DRNN is the noise which mainly includes measurement noise and photon noise. In this section, we test the robustness of the proposed DRNN to photon noise and measurement noise, and show the classification performance as a function of noise level.

The measurement noise is introduced by the sensor in the process of measuring the CS measurement signal level. We simulate various levels of measurement noise by adding zero-mean Gaussian noise to the acquired CS measurements [20], [36]. The noisy measurement vector $y_n$ is generated according to Eq. (5):

$$y_n = y + \hat{y} * r * \epsilon \quad (5)$$

where $\hat{y}$ is the average magnitude of the acquired measurements $y$, the parameter $r$ represents the noise level, defined as the ratio between the standard deviation of added Gaussian noise and $\hat{y}$, and $\epsilon$ is the random noise vector that obeys a multi-dimensional normal distribution.

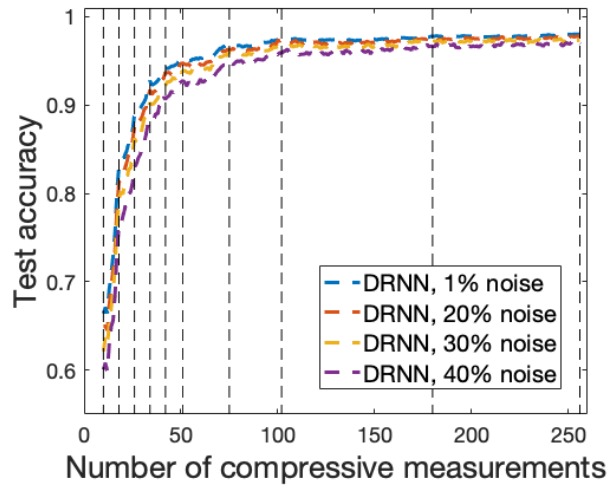

**FIGURE 9.** Test accuracy curves on MNIST for DRNNs under different level of added Gaussian noise. The vertical dash lines indicate the different numbers of measurements used in 10-MR training. The networks are 2L-FFNN with 400 hidden neurons.

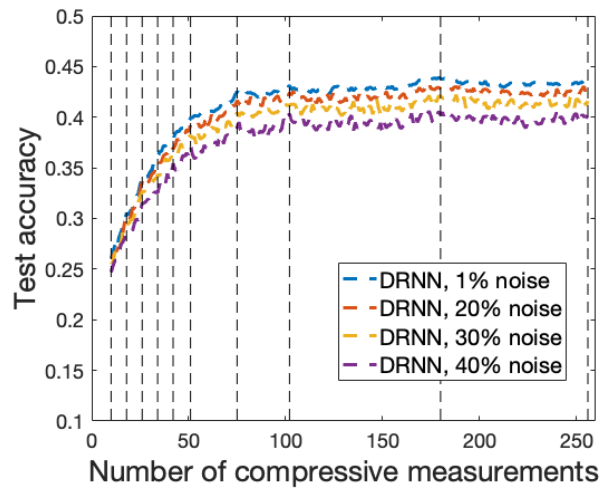

**FIGURE 10.** Test accuracy curves on grayscale CIFAR-10 for DRNNs under different level of added Gaussian noise. The vertical dash lines indicate the different numbers of measurements used in 10-MR training. The networks are 2L-FFNN with 400 hidden neurons.

The photon noise on the image can be modeled by Poisson distribution [37]. Especially in low-light situations, the image could be heavily corrupted by photon noise. The procedure of adding Poisson noise follows the method in [38]. First, the intensity of an 8-bit image is rescaled linearly from range [0, 256] to [0, $n$] at integer levels, where $n$ is a positive integer. Then, for each image pixel, a random number from the Poisson distribution specified by the rate parameter which is the rescaled pixel intensity is generated. This number serves as the pixel intensity for the Poisson noise-added image. Then CS measurement is taken on this image. The value of $n$ is related to the illumination level of the scene. Larger $n$ corresponds to higher illumination level and a cleaner image, whereas smaller $n$ corresponds to lower illumination level where the image is noisier. Example images from Fashion-MNIST and COIL-100 with Poisson noise of $n = 50$ and $n = 200$ are shown in Fig. 3.

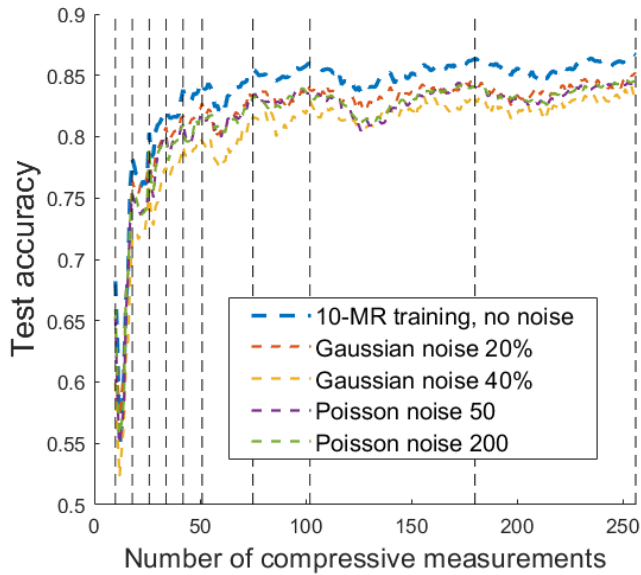

**FIGURE 11.** Test accuracy for DRNNs on Fashion-MNIST dataset with different noise models added. All the DRNNs use 10-MR training with PWH sensing matrices and a CNN with 3 convolutional layers. Comparison are shown among cases of no noise, added Gaussian noise at 20% and 40%, and Poisson noise of $n = 50$ and $n = 200$.

**TABLE 3.** Test accuracy comparison of compressed domain image classification between DRNNs and single-rate neural networks on MNIST dataset.

| | Dynamic-rate 2L-FFNN | | | |
|---|---|---|---|---|
| MR | 1% noise | 20% noise | 30% noise | 40% noise |
| 0.25 | 98.04% | 97.88% | 97.37% | 96.93% |
| 0.1 | 97.46% | 97.34% | 96.64% | 95.87% |
| 0.05 | 95.34% | 94.76% | 93.86% | 92.76% |
| 0.01 | 66.4% | 64.94% | 62.11% | 60.06% |

| | Single-rate 2L-FFNN | | | |
|---|---|---|---|---|
| MR | 1% noise | 20% noise | 30% noise | 40% noise |
| 0.25 | 97.85% | 97.43% | 97.08% | 96.85% |
| 0.1 | 96.69% | 96.59% | 95.85% | 95.25% |
| 0.05 | 95.74% | 95.03% | 94.16% | 92.71% |
| 0.01 | 69.29% | 67.45% | 65.13% | 61.88% |

For noise experiments, we use the same 10-MR training scheme and the same network architectures as in section IV.

**TABLE 4.** Test accuracy comparison of compressed domain image classification between DRNNs and single-rate neural networks on grayscale CIFAR-10.

| | Dynamic-rate 2L-FFNN | | | |
|---|---|---|---|---|
| MR | 1% noise | 20% noise | 30% noise | 40% noise |
| 0.25 | 43.58% | 42.62% | 41.61% | 39.9% |
| 0.1 | 42.68% | 41.63% | 41.10% | 39.29% |
| 0.05 | 39.95% | 39.28% | 38.18% | 37.15% |
| 0.01 | 26.03% | 26.00% | 25.46% | 24.69% |

| | Single-rate 2L-FFNN | | | |
|---|---|---|---|---|
| MR | 1% noise | 20% noise | 30% noise | 40% noise |
| 0.25 | 41.75% | 40.56% | 39.59% | 37.39% |
| 0.1 | 41.11% | 40.28% | 38.63% | 37.37% |
| 0.05 | 37.54% | 37.12% | 35.32% | 34.59% |
| 0.01 | 26.26% | 26.14% | 25.47% | 24.85% |

Training datasets of multiple intermediate MRs and multiple noise levels are generated in the following way: for measurements of each of the 10 MRs, noisy measurements are generated corresponding to $r = \{20\%, 40\%\}$ in the case of Gaussian noise and $n = \{50, 200\}$ in the case of Poisson noise. PWH sensing matrix is used. For illustration purpose, we test the trained DRNNs at Gaussian noise levels of $r = \{1\%, 20\%, 30\%, 40\%\}$ for MNIST and CIFAR-10, as shown in Fig. 9 and Fig. 10, respectively. We also train and test single-rate neural networks at the same Gaussian noise levels and demonstrate results in Table 3 and Table 4. We test at Gaussian noise levels of $r = \{20\%, 40\%\}$ and Poisson noise levels of $n = \{50, 200\}$ for Fashion-MNIST and COIL-100. Results are shown in Fig.11 and Fig. 8, respectively.

We observe that the proposed DRNNs trained with noisy data show the same trend of effectiveness in handling varying MRs and demonstrate good robustness to noise. Also, performance under different noise models and different levels is approximately the same and sometimes better compared to single-rate neural networks.

## VI. CONCLUSION

The proposed DRNN is capable of predicting over a range of MRs on CS measurements acquired by a SPC. The effectiveness of DRNN is demonstrated on datasets of various image scales and using different neural network architectures. The DRNN only needs to be trained on a few intermediate MRs within the range of interest which is efficient in computation and power. The robustness to noise of DRNN is shown under different noise models and noise levels. The CS DRNN method is expected to generalize to other types of random or structured sensing matrix, different neural network architectures, and other compressive sensing related tasks. We feel this approach is not only useful in the single pixel camera architecture but will likely also be beneficial in other compressed sensing camera systems as well.

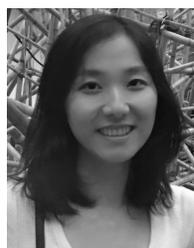

**YIBO XU** received the B.S. degree in physics from Peking University, Beijing, China, in 2013, and the M.S. and Ph.D. degrees in electrical and computer engineering from Rice University, Houston, TX, USA, in 2016 and 2019, respectively. Her current research focuses on computational imaging, computer vision, hyperspectral imaging, and machine learning.

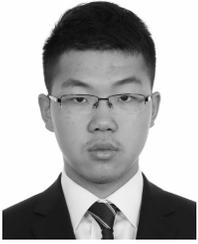

**WEIDI LIU** received the B.S. degree in optical engineering from the University of Rochester, Rochester, NY, USA, in 2018. He is currently pursuing the Ph.D. degree with the Applied Physics Department, Rice University. His current research focuses on computer vision and machine learning.

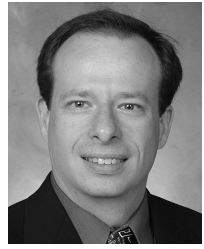

**KEVIN F. KELLY** (Member, IEEE) received the B.S. degree in engineering physics from the Colorado School of Mines, in 1993, and the Ph.D. degree in applied physics from Rice University, in 1999. He is currently an Associate Professor with the Electrical and Computer Engineering Department.

● ● ●